\documentclass[letterpaper, 10 pt, conference]{ieeeconf}  

\IEEEoverridecommandlockouts                              

\usepackage{cite}
\usepackage{amsmath,amssymb,amsfonts}
\usepackage{algorithmic}
\usepackage{graphicx}
\def\BibTeX{{\rm B\kern-.05em{\sc i\kern-.025em b}\kern-.08em
    T\kern-.1667em\lower.7ex\hbox{E}\kern-.125emX}}
\usepackage{textcomp}
\usepackage{xcolor}
\usepackage{booktabs}
\usepackage{url}
\usepackage{float} 
\usepackage{hyperref}
\usepackage{siunitx}

\title{\LARGE \bf
MoiréTac: A Dual-Mode Visuotactile Sensor for Multidimensional Perception Using Moiré Pattern Amplification
}

\author{
Kit-Wa Sou$^{1*}$, Junhao Gong$^{1*}$, Shoujie Li$^{1\dagger}$,\\Chuqiao Lyu$^{1}$, Ziwu Song$^{1}$, Shilong Mu$^{2}$, Wenbo Ding$^{1\dagger}$%
\thanks{*These authors contributed equally to this work.}%
\thanks{$^{\dagger}$Corresponding author: Shoujie Li (lsj20@mails.tsinghua.edu.cn), Wenbo Ding (ding.wenbo@sz.tsinghua.edu.cn)}%
\thanks{$^{1}$Shenzhen Ubiquitous Data Enabling Key Lab, Shenzhen International Graduate School, Tsinghua University, Shenzhen 518055, China.}%
\thanks{$^{2}$Xspark Ai, Shenzhen, China.}
\thanks{This paper has supplementary downloadable material available at: 
\href{https://xiazi718.github.io/MoireTac/}{https://xxx.github.io/MoireTac/}}%
}

\begin{document}

\maketitle
\thispagestyle{empty}
\pagestyle{empty}

\begin{abstract}
Visuotactile sensors typically employ sparse marker arrays that limit spatial resolution and lack clear analytical force-to-image relationships. To solve this problem, we present \textbf{MoiréTac}, a dual-mode sensor that generates dense interference patterns via overlapping micro-gratings within a transparent architecture. When two gratings overlap with misalignment, they create moiré patterns that amplify microscopic deformations. The design preserves optical clarity for vision tasks while producing continuous moiré fields for tactile sensing, enabling simultaneous 6-axis force/torque measurement, contact localization, and visual perception. We combine physics-based features (brightness, phase gradient, orientation, and period) from moiré patterns with deep spatial features. These are mapped to 6-axis force/torque measurements, enabling interpretable regression through end-to-end learning. Experimental results demonstrate three capabilities: force/torque measurement with R²$>$0.98 across tested axes; sensitivity tuning through geometric parameters (threefold gain adjustment); and vision functionality for object classification despite moiré overlay. Finally, we integrate the sensor into a robotic arm for cap removal with coordinated force and torque control, validating its potential for dexterous manipulation.
\end{abstract}


\section{INTRODUCTION}
Robotic manipulation benefits from visuotactile sensing, where cameras observe elastic deformations to reveal texture, contact location, and force cues
(e.g., GelSight~\cite{gelsight}, DIGIT~\cite{digit}, GelSlim~\cite{gelslim}). Yet, most systems rely on sparse marker patterns and black-box inference. Sparse sampling limits information density and differentiability, which hampers 6-axis force/torque estimation and makes calibration and cross-talk analysis difficult in dynamic tasks such as pressing, shearing, and unscrewing~\cite{kappassov2015,shimonomura2019}. In practice, robots also need vision cues before and during contact to localize targets and to keep task context~\cite{li2020}, while tactile cues should quantify forces once contact is made. Bridging these two regimes with a unified sensor and a physically grounded mapping remains challenging.

\begin{figure}[t]
    \centering
    \includegraphics[width=\columnwidth]{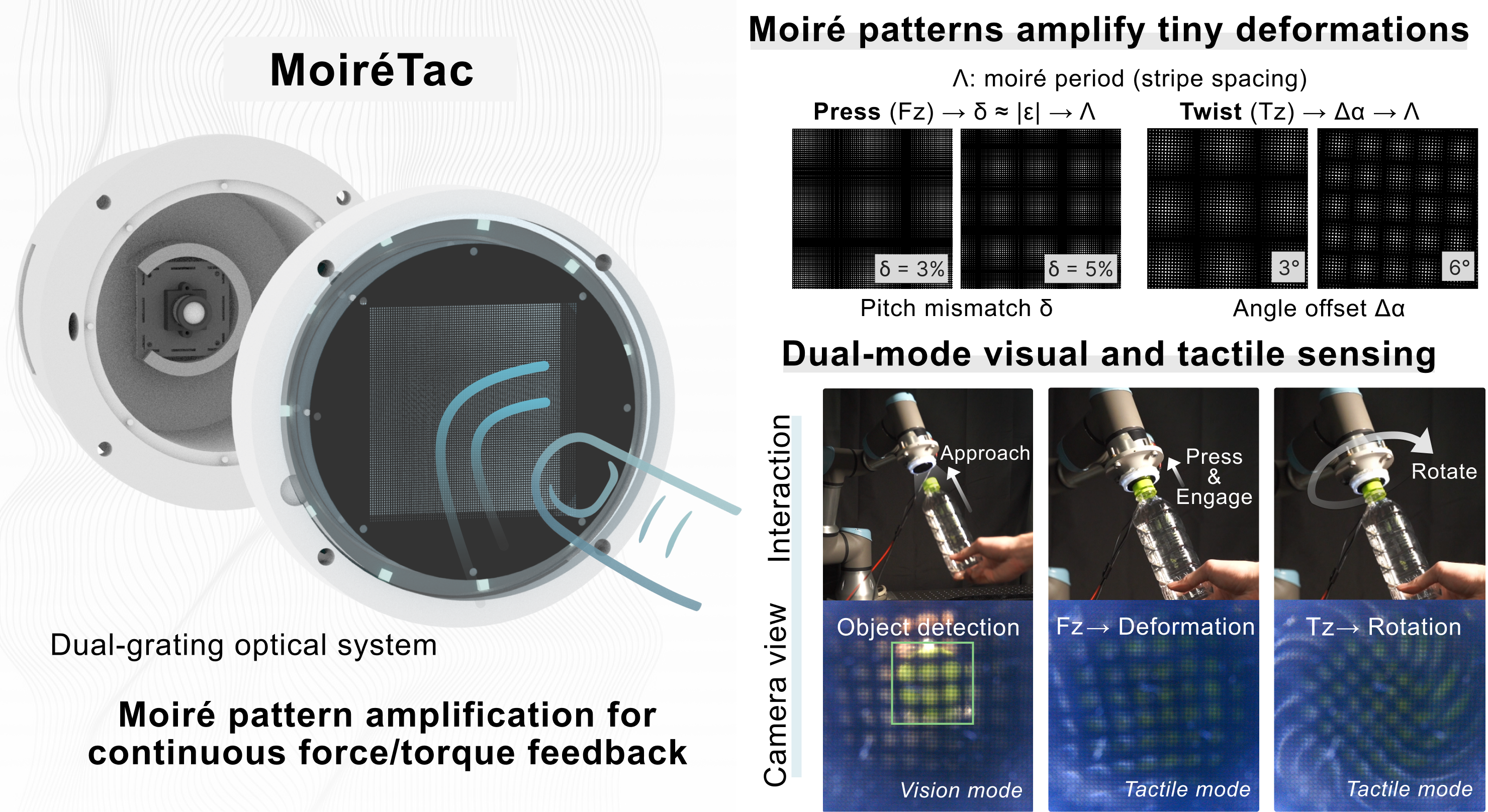} 
    \caption{Overview of MoiréTac. The dual-grating optical system generates moiré observables, which are mapped to force/torque measurements. The top-left panels show how press and rotation affect stripe density and orientation. The application demonstrates dual-mode visual and tactile sensing during robotic manipulation.}

    \label{fig:Overview}
\end{figure}

Moiré patterns, formed by the interference of two slightly misaligned periodic structures, offer a dense, continuous texture that is sensitive to microscopic deformations \cite{Moire_formation}. Compared to sparse marker patterns, which are often limited in their spatial resolution and can create challenges in force-to-image mapping \cite{Moire_sparse_vs_dense}, moiré-based textures provide rich, high-density information that can be directly related to physical deformations. This makes them highly effective for force/torque estimation in dynamic tasks, where continuous measurement is crucial \cite{Moire_torque_estimation}. Furthermore, the continuous nature of moiré patterns allows for more accurate calibration and cross-talk analysis, particularly in tasks involving pressing, shearing, and unscrewing \cite{Moire_calibration}. By leveraging these high-resolution, interpretable patterns, visuotactile sensors can offer both vision cues for target localization and tactile feedback for force regulation, addressing the challenges of multimodal sensing without gaps in performance \cite{Moire_multimodal_sensing}.

In this paper, we introduce \textbf{MoiréTac}, a compact visuotactile sensor that converts small deformations into a dense moiré field using a double-grid optic (Fig.~\ref{fig:Overview}). Our study presents three major contributions:

\begin{itemize}
  \item \textbf{Moiré interferometry for visuotactile sensing:} We apply moiré pattern amplification to transform marker-based sensing into continuous interference fields, achieving \(R^2 > 0.98\) across six force/torque axes. Unlike sparse marker methods that interpolate between discrete points, our continuous fields directly encode mechanical deformations at every pixel.

  \item \textbf{Physics-grounded force/torque framework:} We establish analytical mappings from four moiré observables (\(I\), \(\nabla\phi\), \(\theta\), \(\Lambda\)) to 6-axis wrench, enabling sensitivity adjustment through geometric parameters. This interpretable approach permits analytical sensitivity analysis and systematic calibration transfer, capabilities absent in pure learning-based methods.

  \item \textbf{Transparent dual-mode operation:} We maintain optical transparency for object recognition while measuring forces, demonstrated in automated cap removal tasks. The preserved visual channel operates concurrently with tactile sensing, unlike opaque sensors that lose visual information upon contact.
\end{itemize}

\section{RELATED WORK}

In visuotactile sensing, an onboard camera images an elastic medium under contact, converting deformations into spatial measurements (e.g., GelSight~\cite{gelsight}, DIGIT~\cite{digit}, GelSlim~\cite{gelslim})\cite{DingJSTSP2024,VBTSClassify2025}. However, comprehensive reviews\cite{luo2017,dahiya2010,kappassov2015,shimonomura2019,DingJSTSP2024,VBTSClassify2025} identify two persistent limitations: sparse marker arrays that undersample contact regions, and black-box force mappings lacking physical interpretability. These constraints complicate 6-axis force/torque estimation and cross-device calibration transfer.

Recent systems~\cite{AllSight,GelSight360,9DTact,BioTacTip,ViTacTip,ShadowTac} employ deep networks to directly regress forces from tactile images, achieving strong accuracy in controlled settings. Yet these data-driven approaches struggle with generalization across devices and require extensive calibration~\cite{EasyCalib,FeelAnyForce}. The absence of physics-grounded representations impacts multi-axis decoupling, as pure learning methods cannot leverage analytical relationships between image features and mechanical states.

Moir\'e interferometry offers an alternative via optical amplification of microscopic deformations~\cite{Moire_formation}. Unlike discrete markers, moir\'e fringes produce continuous fields whose period, orientation, and phase relate analytically to mechanical quantities~\cite{interferometry}, enabling sub-micron precision in structural monitoring and nano-positioning. For tactile use, traditional moir\'e force visualizers remain qualitative and parallax-prone; fiber–optic plates mitigate parallax yet still prioritize visual readout over dense 6-DoF estimation~\cite{Zuo2025}. Meanwhile, Moir\'eVision~\cite{MoireVision} demonstrated 6-DoF tracking with moir\'e cues, suggesting that calibrated moir\'e parameters can yield dense, interpretable measurements for advanced tactile sensing, including force/torque inference~\cite{Moire_calibration,Moire_torque_estimation}.

Multimodal designs combining vision and touch typically require separate optical paths or complex multiplexing~\cite{CompdVision,StereoTac,TIRgel,JamTac,ICRA2023,TRO2024}, increasing system complexity. Transparent sensors could preserve visual channels through the tactile medium, yet existing approaches sacrifice either optical clarity or tactile resolution~\cite{TIRgel,CompdVision}.

MoiréTac addresses these gaps by: (1) replacing sparse markers with continuous moiré fields for dense spatial measurements, (2) establishing analytical mappings from four moiré observables to 6-axis forces/torques, and (3) maintaining optical transparency for dual-mode vision-tactile operation. This physics-informed approach bridges precision measurement with learning-based adaptability.
\begin{figure*}[ht]
  \centering
  \includegraphics[width=1.9\columnwidth]{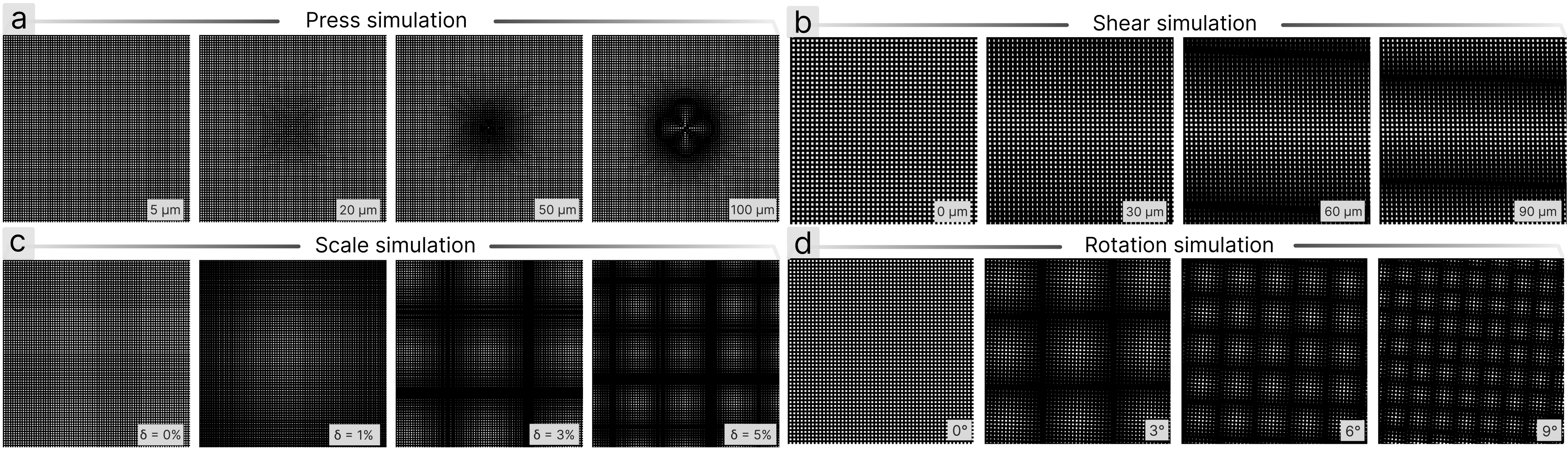}
  \caption{Simulation panels: 
  (a) Press simulation with displacements of 5 \si{\micro\meter} to 100 \si{\micro\meter}. 
  (b) Shear simulation with displacements from 0 \si{\micro\meter} to 90 \si{\micro\meter}. 
  (c) Scale simulation with scaling variations from 0\% to 5\%. 
  (d) Rotation simulation with angular shifts from 0° to 9°. 
  All displacements and rotations represent the changes in the moiré patterns, used to simulate different types of mechanical deformations.}
  \label{fig:sim}
\end{figure*}

\section{DESIGN AND IMPLEMENTATION}

MoiréTac addresses the sparse sampling and weak physical grounding of current visuotactile sensors through moiré interferometry. The continuous interference patterns provide dense, differentiable signals with established optical theory linking image features to mechanical loads. 

\subsection{Design Principle}\label{sec:design-principle}

MoiréTac exploits moiré interferometry to transform microscopic deformations into 
rich visual signatures. Two stacked micro-gratings create an optical system where 
mechanical changes produce dramatic fringe variations, naturally encoding the 
complete 6-axis wrench through four observables: intensity \(I\), phase gradient \(\nabla\phi\), 
fringe orientation \(\theta\), and period \(\Lambda\).

\textbf{Mathematical Foundation:} Each grating functions as a spatial harmonic filter, and modulates the intensity distribution in space. The grating intensity distribution is described by:
\begin{equation}
g_i(\mathbf{x})=\cos(\mathbf{k}_i\!\cdot\!\mathbf{x}),\quad
\mathbf{k}_i=\tfrac{2\pi}{p_i}[\cos\alpha_i,\sin\alpha_i]^\top,
\label{eq:grating}
\end{equation}
where $p_i$ and $\alpha_i$ define pitch and orientation. When overlapped, they generate a beat pattern with moiré wavevector

\vspace{-4mm}
\begin{equation}
\mathbf{K} = \mathbf{k}_1 - \mathbf{k}_2, \quad \Lambda = \frac{2\pi}{\|\mathbf{K}\|}, \quad \theta = \operatorname{atan2}(K_y, K_x).
\label{eq:moireK}
\end{equation}
This immediately provides us with the period \(\Lambda\) and orientation \(\theta\) as directly measurable observables.

\textbf{Normal Force via Optical Coupling (Fig.~\ref{fig:sim}a):} When a hemispherical indenter (mimicking fingertip contact) is pressed onto the surface, Poisson contraction induces lateral strain \( \varepsilon \), which modulates the pitch mismatch and thus the moiré period, encoding the normal force.

\vspace{-4mm}
\begin{equation}
I(\mathbf{x}) \!\approx\! I_0+\kappa\,(P\!\ast\! h)(\mathbf{x}),\quad
\Lambda' \approx \frac{p}{\delta'},\;\;\delta'\!\approx\!|\varepsilon(F_z)|,
\label{eq:intensity-press}
\end{equation}
where $P\!\ast\! h$ represents the pressure field convolved with the optical point spread function. This dual response makes $F_z$ particularly robust; note that the net fringe trend under compression is set by Eq.~\eqref{eq:deff}.

\textbf{Shear Forces via Phase Tracking (Fig.~\ref{fig:sim}b):} The beauty of moiré lies in its sensitivity to lateral motion. Any local displacement $\mathbf{u}(\mathbf{x})$ directly modulates the fringe phase:
\begin{equation}
\phi(\mathbf{x}) \approx \mathbf{K}\!\cdot\!\mathbf{x} + \mathbf{K}\!\cdot\!\mathbf{u}(\mathbf{x}).
\label{eq:phase}
\end{equation}
The spatial averages of phase gradients $\langle\partial_x\phi\rangle$ and $\langle\partial_y\phi\rangle$ elegantly encode the net shear forces $F_x$ and $F_y$, providing precise displacement sensitivity. While physical contact requires normal preload to generate shear, our simulation isolates the lateral response to clearly demonstrate phase modulation under pure shear loading.

\textbf{Geometric Design Space (Fig.~\ref{fig:sim}c):} The sensor's sensitivity is tunable through geometric parameters. The general period relationship:
\begin{equation}
\Lambda=\frac{p_1 p_2}{\sqrt{\,p_1^{2}+p_2^{2}-2 p_1 p_2 \cos\Delta\alpha\,}}
\label{eq:general-lambda}
\end{equation}
simplifies to two practical design regimes:
\begin{equation}
\Lambda\approx\frac{p}{\Delta\alpha}\quad\text{(angle-dominated)},\quad
\Lambda=\frac{p}{\delta}\quad\text{(pitch-dominated)},
\label{eq:angle-pitch}
\end{equation}
where \(p=(p_1{+}p_2)/2\) and \(\delta=|p_2{-}p_1|/p\). Smaller \(\Delta\alpha\) or \(\delta\) yields larger \(\Lambda\) (greater geometric magnification \(A\)). 

\vspace{-2mm}
\begin{equation}
\delta_{\mathrm{eff}}\approx\big|\delta_{\mathrm{obj}}-a/Z\big|,
\label{eq:deff}
\end{equation}
with inter-grating spacing \(a\) and camera distance \(Z\).
Under compression, the sign of \(\delta_{\mathrm{obj}}-a/Z\) sets the trend:
if \(\delta_{\mathrm{obj}}<a/Z\), then \(\delta_{\mathrm{eff}}\) decreases and \(\Lambda\) increases (sparser);
if \(\delta_{\mathrm{obj}}>a/Z\), the opposite holds (denser).

\textbf{Moment Sensing via Pattern Morphology (Fig.~\ref{fig:sim}d):} Torques manifest through distinct fringe transformations. Twist $T_z$ simply rotates the entire pattern:
\vspace{-2mm}
\begin{equation}
\Delta\theta \approx \Delta\alpha \;\propto\; T_z.
\label{eq:twist}
\end{equation}

Tilting moments \(T_x, T_y\) create asymmetric pressure distributions, shifting the brightness centroid \(\mathbf{c} = \langle \mathbf{x}\rangle_I\):

\vspace{-1mm}
\begin{equation}
[T_x,T_y]^\top \approx \mathbf{M}\,(\mathbf{c}-\mathbf{c}_0),
\label{eq:tilt}
\end{equation}
where $\mathbf{M}$ is a calibrated mapping matrix. Table~\ref{tab:quickmap} summarizes these physics-guided relationships, forming our compact feature pipeline. These analytical relationships guide the sensor's physical implementation, where geometric parameters and material properties must be carefully orchestrated.

\begin{table}[ht!]
\centering
\caption{Mapping of moiré observables to force/torque measurements}
\label{tab:quickmap}
\begin{tabular}{p{0.07\linewidth} p{0.83\linewidth}}
\toprule
\textbf{Axis} & \textbf{Primary observable \& physical mechanism}\\
\midrule
$F_z$ & Brightness $I$ rises with contact; period $\Lambda$ varies via strain coupling \text{Eq.}~\eqref{eq:intensity-press}\\
$F_x,F_y$ & Phase gradients $\langle\nabla\phi\rangle$ track lateral displacement fields \text{Eq.}~\eqref{eq:phase}\\
$T_z$ & Orientation $\theta$ rotates directly with applied twist \text{Eq.}~\eqref{eq:twist}\\
$T_x,T_y$ & Brightness centroid $\mathbf{c}$ shifts under asymmetric loading \text{Eq.}~\eqref{eq:tilt}\\
\bottomrule
\end{tabular}
\end{table}

\begin{figure}[ht!]
    \centering
    \includegraphics[width=\columnwidth]{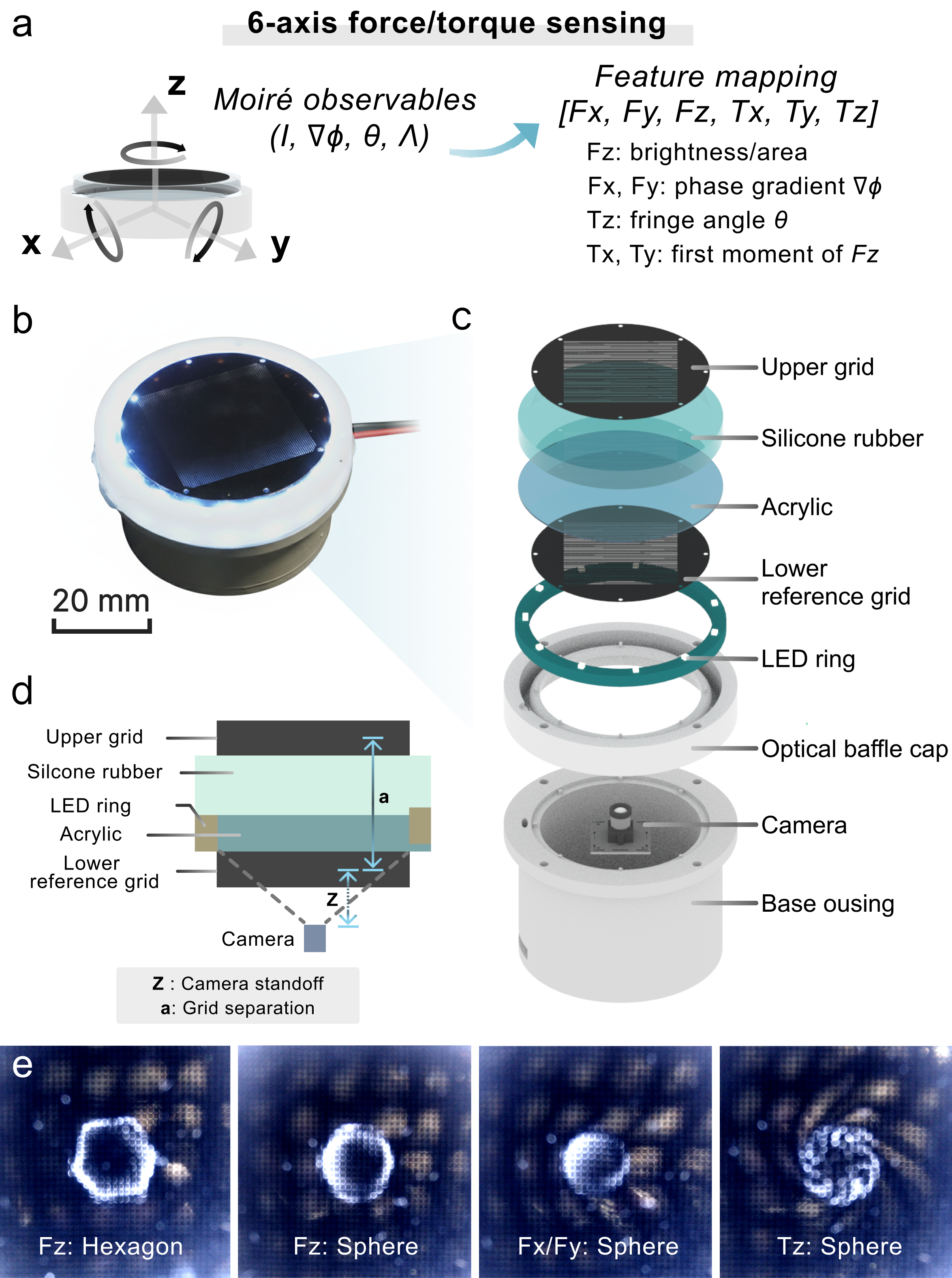}
    \caption{Structure illustration. (a) Mapping from Moiré observables (intensity \(I\), phase \(\nabla\phi\), angle \(\theta\), period \(\Lambda\)) to 6-axis force/torque sensing. (b) Photo of prototype. (c) Exploded view of layered architecture. (d) Cross-section showing compression–to–fringe coupling. (e) Responses under normal, shear, and twist loading; a waveguided LED produces a contact rim that delineates the boundary and maintains fringe visibility.}

    \label{fig:structure}
\end{figure}

\begin{figure*}[ht]
    \centering
    \includegraphics[width=1.55\columnwidth]{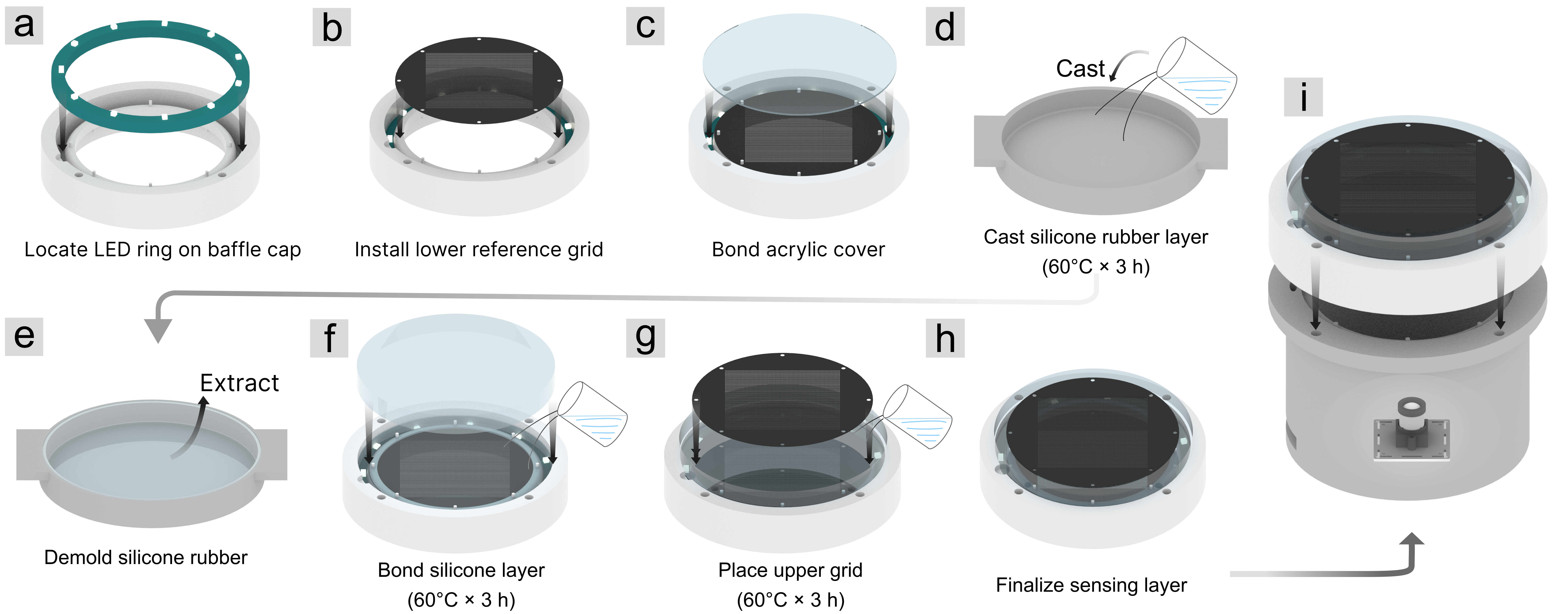}
    \caption{Fabrication process of MoiréTac sensor showing assembly steps. (a-c) Optical base assembly with LED ring positioning on baffle cap, lower reference grid installation, and acrylic cover bonding. (d-f) Elastomer layer preparation including silicone casting at 60°C for 3 hours, demolding, and thermal bonding to acrylic substrate. (g-h) Sensing layer completion with upper grid placement and final assembly. (i) Fully integrated sensor with camera.}
    \label{fig:fabrication}
\end{figure*}

\subsection{Mechanical Stack and Structure}\label{sec:stack}

To realize these optical principles (Fig.~\ref{fig:structure}a), MoiréTac employs a compact layered architecture that transforms microscopic deformations into vivid moiré patterns (Fig.~\ref{fig:structure}b). The sensor achieves 4–24× mechanical-to-optical amplification by exploiting interference between two micro-gratings separated by a compliant medium. The layered architecture (Fig.~\ref{fig:structure}c) integrates optical and mechanical functions: a deformable upper grating moves with applied loads, while a fixed lower reference grating provides the interferometric baseline; a silicone elastomer bonded to an acrylic carrier offers compliance and transparency; an LED ring with a shallow baffle ensures uniform illumination; and a bottom camera captures the resulting fringes.

The key mechanism lies in the inter-grating spacing \(a\) (Fig.~\ref{fig:structure}d). Under compression, \(a\) decreases and induces Poisson-driven lateral strain, which modulates the pitch mismatch \(\delta\) and hence the moiré period \(\Lambda\), encoding the normal force. Lateral shifts produce fringe translations that encode shear forces. An LED ring waveguided through PDMS forms a bright contact rim whose integrated brightness/area rises with compression, thereby localizing contact (Fig.~\ref{fig:structure}e). This photometric rim provides a visual boundary cue that is distinct from the tactile moiré features. The same illumination mitigates fringe visibility loss in the dark annulus under stronger compression. Finally, we preset a small \(\delta\) to ensure robust phase extraction (Sec.~\ref{sec:design}).


\subsection{Fabrication}
MoiréTac assembly follows an eight-step procedure (Fig.~\ref{fig:fabrication}); Table~\ref{tab:bom} lists components for the grating set used in subsequent experiments. The optical base assembly (a--c) begins with positioning the LED ring on the baffle cap, installing the lower reference grid at the focal plane, and bonding an acrylic cover using cyanoacrylate adhesive to create a flat carrier platform.

The sensing layer (d--h) uses PDMS elastomer (\(30{:}1\) ratio) cast onto the acrylic substrate and cured at \(60\,^{\circ}\mathrm{C}\) for 3~hours. After demolding, the PDMS is thermally bonded to the acrylic. The upper grid is positioned with a controlled pitch difference \(\delta\) relative to the lower grid, which generates magnified moiré patterns. Critical parameters include grid alignment (\(\pm 0.5^{\circ}\)) and uniform elastomer thickness.

\begin{table}[!ht]
\centering
\caption{Bill of materials for MoiréTac sensor}
\label{tab:bom}
\begin{tabular}{l|l|l}
\toprule
\textbf{Component} & \textbf{Description} & \textbf{Process} \\
\midrule
\multicolumn{3}{l}{\textit{Optical Components}} \\
Camera & IMX258, 12MP, 120° FOV & Off-the-shelf \\
Upper grid & 350 \si{\micro\meter} pitch on black NBR & Laser ablation \\
Lower reference grid & 300 \si{\micro\meter} pitch on black NBR & Laser ablation \\
LED ring & 40 mm dia., white LEDs & Off-the-shelf \\
\midrule
\multicolumn{3}{l}{\textit{Structural Components}} \\
Base housing & PLA, 45 mm dia. & 3D printing \\
Optical baffle cap & PLA, 40 mm dia. & 3D printing \\
Camera mount & PLA, M12 thread & 3D printing \\
\midrule
\multicolumn{3}{l}{\textit{Elastomer Assembly}} \\
Acrylic & 1 mm clear sheet & Laser cutting \\
Silicone rubber & PDMS 30:1, 3 mm thick & Molding \\
Adhesive & Cyanoacrylate 502 & Manual \\
\midrule
\multicolumn{3}{l}{\textit{Raw Materials}} \\
Nitrile rubber sheet & Black NBR, 0.5 mm thick & Die cutting \\
\bottomrule
\end{tabular}
\end{table}

\begin{figure}[!ht]
    \centering
    \includegraphics[width=\columnwidth]{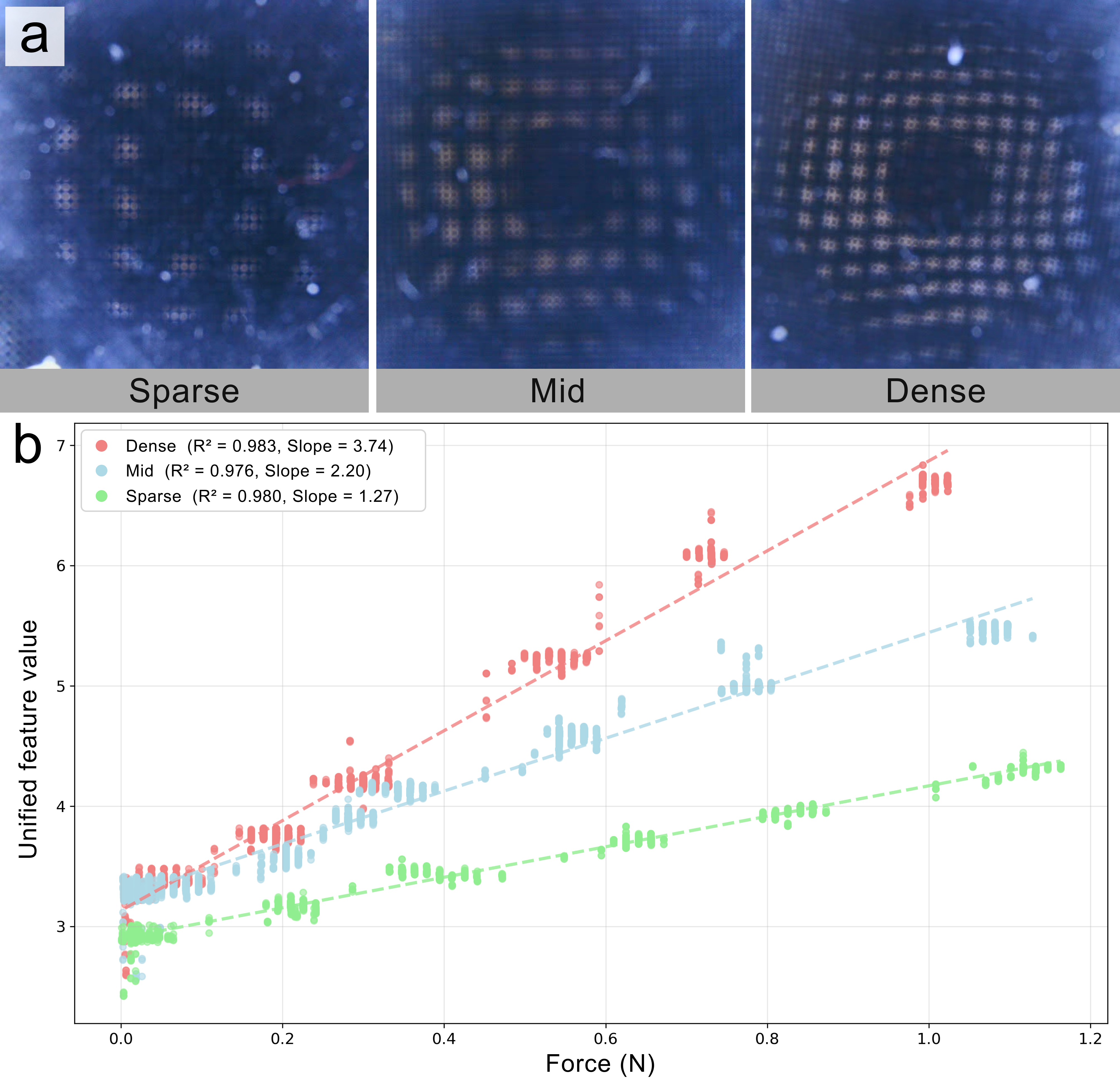}
    \caption{Sensitivity modulation via pitch difference \(\delta\). (a) Moiré patterns 
    under three design configurations showing increasing fringe density, with the central dark circle indicating the robot arm's shadow.
    (b) Force–response curves demonstrating tunable sensitivity.}
    \label{fig:sensitivity}
\end{figure}

\begin{figure}[!h]
    \centering
    \includegraphics[width=\columnwidth]{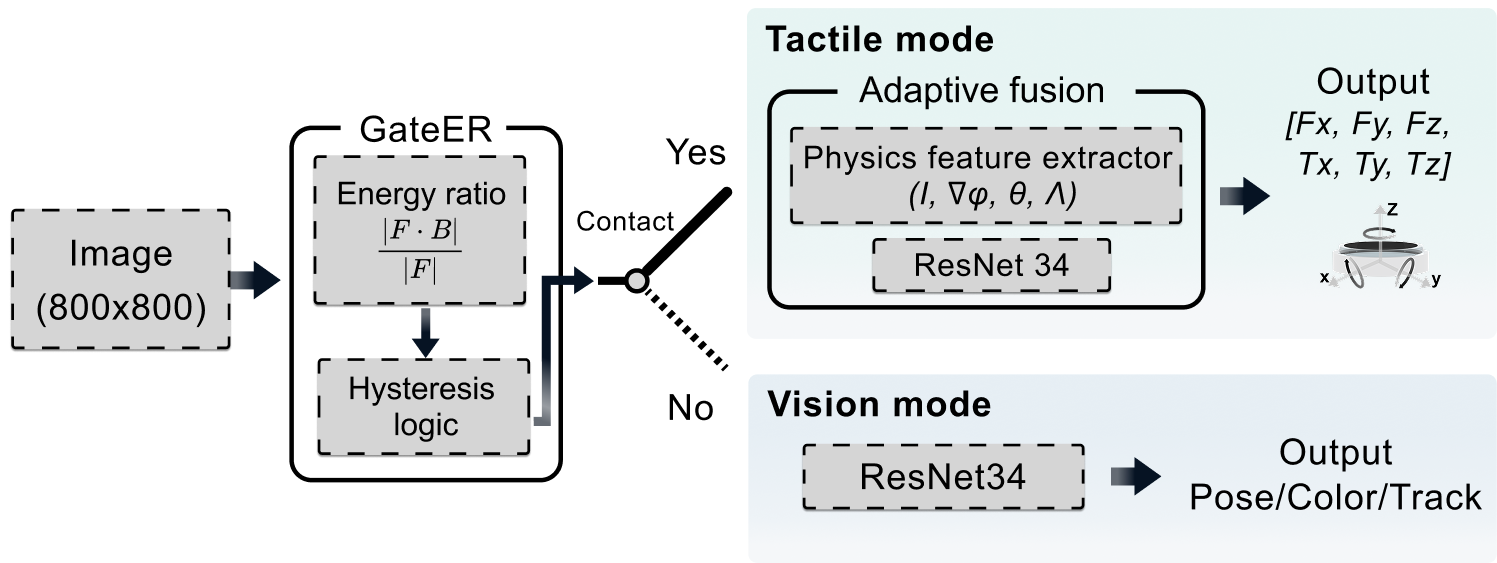}
    \caption{Overview of the MoiréTac processing pipeline.}
    \label{fig:pipeline}
\end{figure}

\begin{figure*}[ht]
    \centering
    \includegraphics[width=1.65\columnwidth]{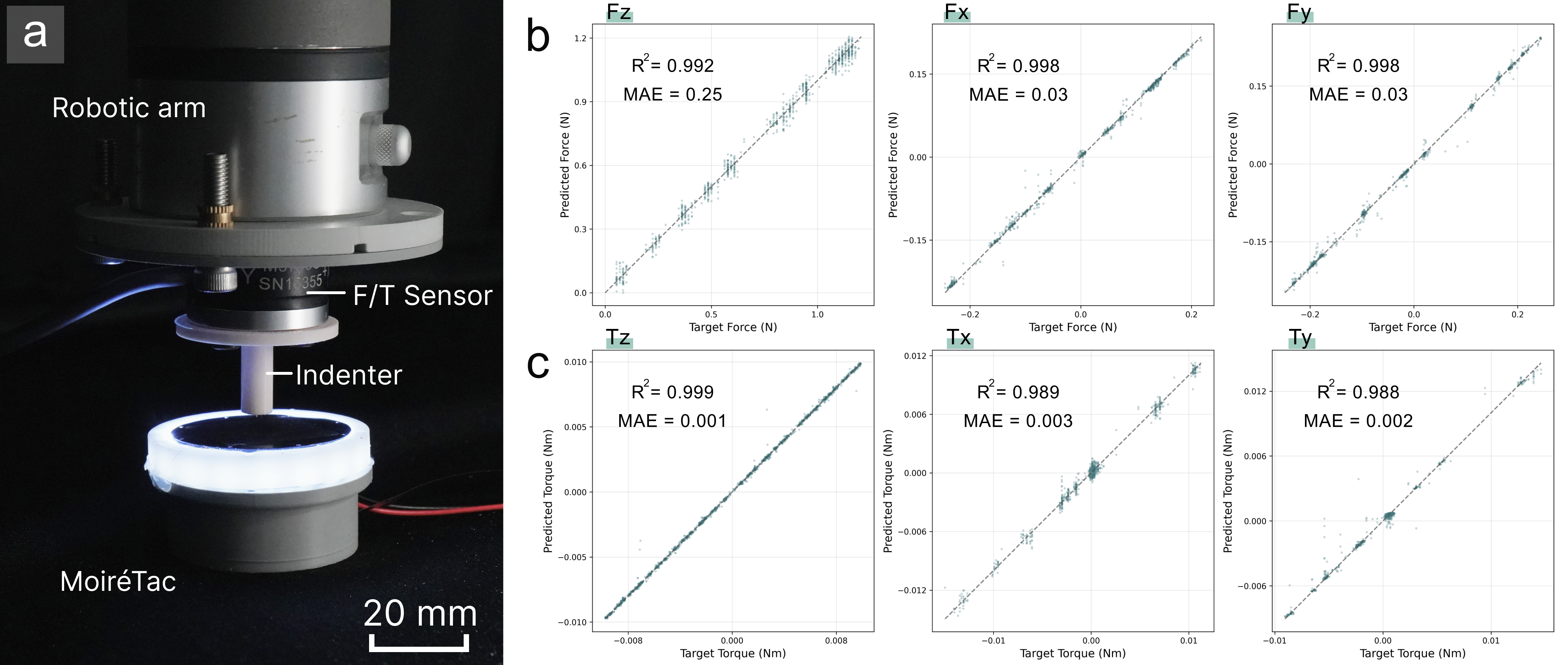}
    \caption{Six-axis force/torque calibration. (a) Experimental setup with robotic arm, commercial F/T reference, and MoiréTac prototype. (b) Force calibration showing linearity for normal force \(F_z\) and shear forces \(F_x, F_y\). (c) Torque calibration demonstrating \(T_z\) and \(T_x, T_y\) performance.}
    \label{fig:calibration}
\end{figure*}

\subsection{Sensitivity Tuning through Geometric Design}\label{sec:design}

Moir\'eTac’s phase sensitivity is governed by the effective pitch mismatch $\delta_{\mathrm{eff}}$ (Eq.~\ref{eq:deff}), which combines the intrinsic mismatch $\delta_{\mathrm{obj}}$ with the fixed depth separation, so larger $\delta_{\mathrm{eff}}$ shortens the moir\'e period $\Lambda$ and increases fringe density, thereby enhancing sensitivity. All three configurations satisfy \(\delta_{\mathrm{obj}}<a/Z\), so compression increases \(\Lambda\) (sparser). We evaluate three pitch pairs (near grid at $Z$ with pitch $p_2$; far grid at $Z+a$ with pitch $p_1$; $a/Z = 0.25$), matching the panel tags in Fig.~\ref{fig:sensitivity}a: Dense $(p_1,p_2)=(0.20,0.20)\,\mathrm{mm}$ gives $A=4.00$ and $\delta_{\mathrm{eff}}=0.25$; Mid $(0.35,0.30)$ gives $A=14.00$ and $\delta_{\mathrm{eff}}=0.10$; Sparse $(0.30,0.25)$ gives $A=24.00$ and $\delta_{\mathrm{eff}}=0.07$. As fringe density decreases from Dense to Sparse (Fig.~\ref{fig:sensitivity}b), sensitivity decreases accordingly; note that $A$ increases, but the readout is governed by fringe density. Selecting $\delta$ at fabrication thus tunes the sensitivity–range trade-off: denser fringes increase sensitivity but narrow range; sparser widen range but reduce sensitivity. We adopt Mid for balance.

\subsection{Sensing Pipeline}

\textbf{Image Preprocessing.} Although the 120° lens introduces minimal distortion, we perform a one-time calibration to establish the pixel-to-millimeter mapping and define the 30×30 mm sensing region (Fig.~\ref{fig:pipeline}). A cylinder array pressed onto the sensor creates reference points for cropping and scaling the raw image to a standardized 800×800 pixel format (20 pixels/mm), ensuring consistent input dimensions for the neural network and preserving spatial alignment.

\textbf{Dual-Mode Processing.} The system employs a GateER (Gated Energy Ratio) module, which computes an energy metric from the moiré pattern to detect contact. The module uses an adaptive threshold mechanism with approximately 20\% hysteresis margin to prevent mode oscillation, achieving typical switching response times of 30-40 ms. The hysteresis logic settings need to be tuned based on specific application requirements. In vision mode (no contact), ResNet34 processes the undistorted moiré pattern as a textured background for pose estimation, color recognition, and object tracking. Upon contact detection, the system switches to tactile mode, activating an adaptive fusion mechanism that interprets fringe deformations as mechanical signals for real-time force sensing at camera frame rate.

\textbf{Force/Torque Regression.} Upon switching to tactile mode, MoiréTac employs an adaptive fusion architecture that processes calibrated images through dual parallel pathways. The physics feature extractor computes four primary observables, brightness \(I\), phase gradients \(\nabla\phi\), orientation \(\theta\), and period \(\Lambda\), based on moiré theory relationships (Table~\ref{tab:quickmap}), while ResNet34 independently extracts deep spatial features. These complementary representations merge before the final regression layers, combining physics-based constraints with data-driven pattern recognition. This hybrid approach leverages the interpretability of moiré theory while allowing the network to learn complex nonlinearities and cross-axis couplings beyond simplified analytical models. The network outputs a 6-dimensional vector \([F_x, F_y, F_z, T_x, T_y, T_z]\) through modified fully connected layers. ResNet34's residual connections work well for moiré patterns, preserving both fine-grained local variations and global fringe structures in the architecture.

\section{Experimental Evaluation}

\subsection{Force/Torque Characterization}\label{sec:calibration}
We validate MoiréTac against a commercial F/T sensor (Fig.~\ref{fig:calibration}a) under laboratory white light conditions. A 6-DoF robot applies controlled loads while both sensors are recorded synchronously at 60\,Hz; frames and F/T readings are time-aligned and downsampled to the camera rate. We collected approximately 10,000 image-force pairs over multiple sessions, with uniform sampling across each axis's operating range. Each axis is calibrated in isolation to reduce coupling: (i) normal indentations for \(F_z\) (0–1.2\,N); (ii) lateral sliding for \(F_x,F_y\) ($\pm$0.2 N) under a constant preload; (iii) axial rotation about the sensor center for \(T_z\) ($\pm$0.008 Nm); and (iv) off-center indentations at multiple locations for \(T_x,T_y\) ($\pm$0.012 Nm). Rectified \(800\times800\) grayscale images are fed to a physics-informed ResNet-34 for regression. The dataset was split 80/20 for training and testing, with stratification across load levels and contact locations.

Results demonstrate agreement across all six axes (Fig.~\ref{fig:calibration}b,c). Forces achieve \(R^2 \ge 0.99\) (.\(F_z\): \(R^2=0.992\), MAE \(=0.25\,\mathrm{N}\)). Torque estimation shows \(T_z\) with the strong accuracy due to direct fringe-rotation coupling (\(R^2=0.999\), MAE \(=1\times10^{-3}\,\mathrm{Nm}\)), while tilt moments \(T_x, T_y\) maintain \(R^2>0.98\). The tight clustering along identity lines indicates low cross-talk between axes under isolated loading conditions on the prototype.

    \label{fig:ablation}


\subsection{Visual Sensing}\label{sec:vision}

Unlike opaque tactile sensors that occlude the contact area, MoiréTac's transparent dual-grating architecture preserves the visual channel. Although moiré fringes overlay the scene, they behave as high-frequency spatial modulation that can be filtered or ignored by recognition algorithms. This transparency enables pre-contact identification and alignment, which are critical when grasping unfamiliar objects or targeting specific surface features.

\begin{figure}[ht]
    \centering
    \includegraphics[width=1.\columnwidth]{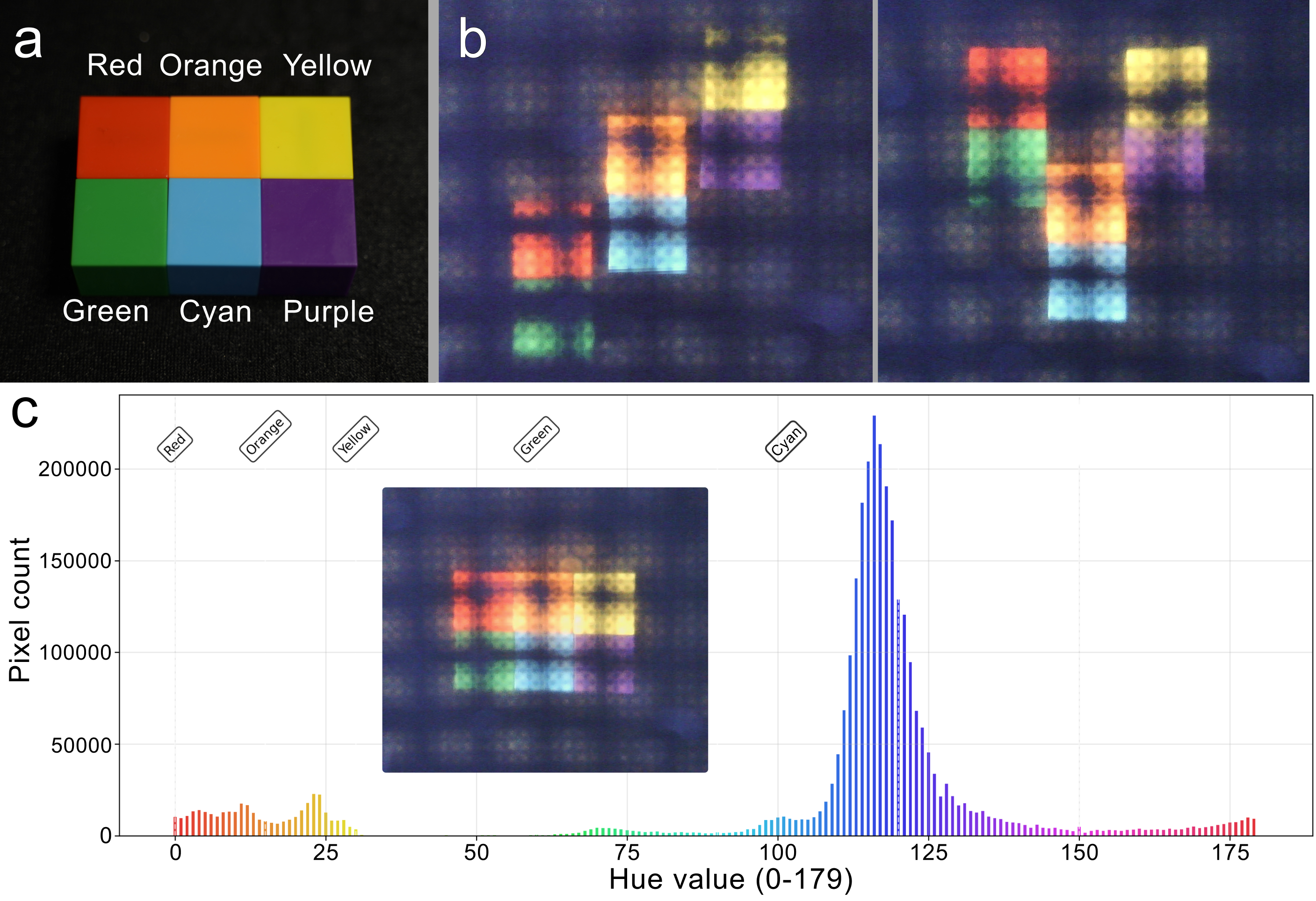} 
    \caption{Color perception in vision mode. (a) Reference cube with six colors. (b) Images captured under two poses; the colored surface is visible on top of the moiré background. (c) Hue histogram in HSV (OpenCV scale 0--179) aggregated over the patches; distinct peaks for red, orange, yellow, green, cyan, and purple indicate separability for identification.}
    \label{fig:vision_color}
\end{figure}

We validate color fidelity using a standard six-color card (Fig.~\ref{fig:vision_color}a). Despite transmission through two micro-gratings and an elastomer layer, colors remain clearly distinguishable (Fig.~\ref{fig:vision_color}b). The HSV hue histogram (Fig.~\ref{fig:vision_color}c) shows well-separated peaks with limited overlap between adjacent classes. The moiré pattern introduces periodic intensity modulation yet preserves underlying spectral content, enabling effective color-based segmentation.

\begin{figure}[ht]
    \centering
    \includegraphics[width=1.\columnwidth]{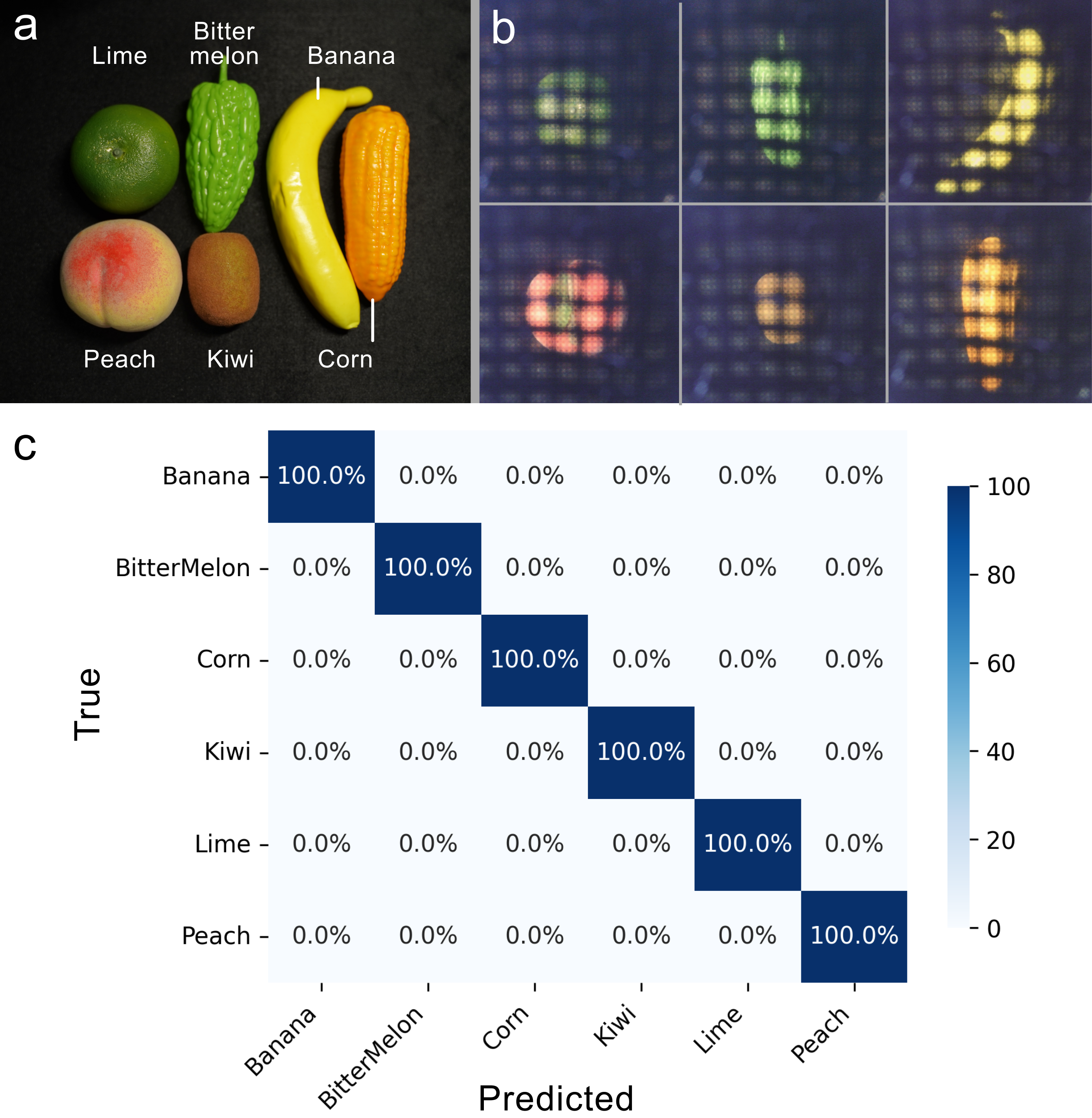} 
    \caption{Vision-mode object recognition. (a) Six target fruits. (b) Representative sensor images where color appearance is visible over the moiré background. (c) Confusion matrix of trained classifier on test set}
    \label{fig:vision_fruit}
\end{figure}

Object recognition tests with six fruits/vegetables demonstrate practicality (Fig.~\ref{fig:vision_fruit}a). We collected 900 samples for each of six fruits in 3 different orientations at 60\,Hz, capturing 5 seconds per orientation. The dataset was divided using an 80/20 train-test split. Distinct color–texture cues remain visible through the sensor (Fig.~\ref{fig:vision_fruit}b): e.g., lime's uniform green, bitter melon's ridges, banana's smooth yellow curve, peach's gradient, kiwi's brown fuzz, and corn's kernel regularity. A ResNet34 classifier achieves high accuracy (Fig.~\ref{fig:vision_fruit}c), reaching 100\% accuracy on our test set without requiring specialized preprocessing to remove moiré artifacts.


\begin{figure}[ht]
    \centering
    \includegraphics[width=1\columnwidth]{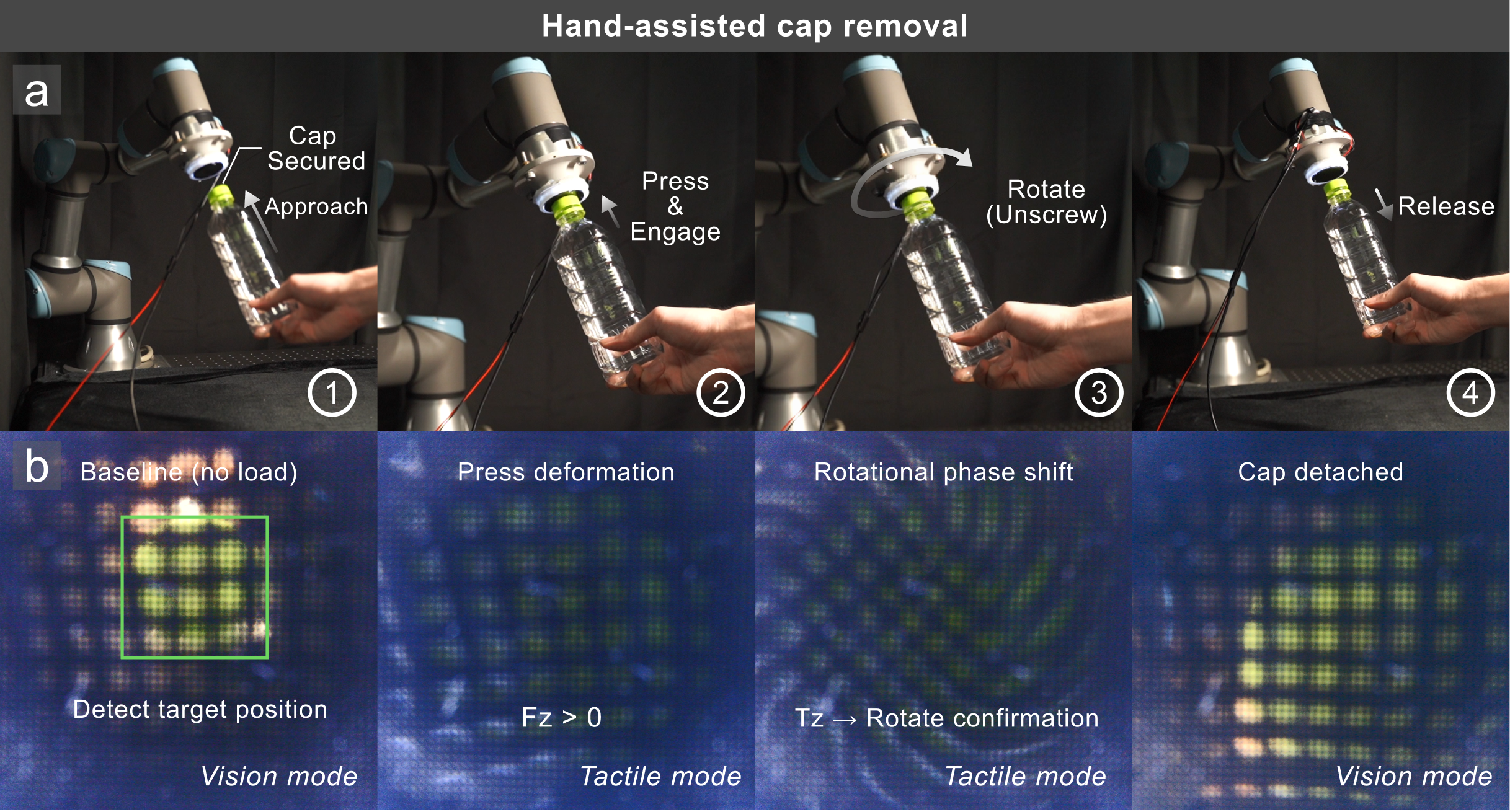}
    \caption{Hand-assisted cap removal demonstrating dual-mode operation. (a) Task sequence: (1) vision-guided approach to locate cap, (2) press to establish grip with force feedback, (3) rotate to unscrew with torque monitoring, (4) release after completed removal. (b) Corresponding sensor views showing vision-tactile mode transitions. Green box indicates visual target detection, while moiré patterns reveal force/torque during manipulation.}
    \label{fig:cap_removal}
\end{figure}

\subsection{Dual-Mode Manipulation Tasks}

To demonstrate MoiréTac's vision-tactile integration, we evaluate the sensor on a bottle cap removal task requiring both modalities and real-time feedback (Fig.~\ref{fig:cap_removal}a). This manipulation sequence showcases automatic mode switching based on contact state and highlights the value of preserving visual feedback throughout the task.

The task proceeds through four phases with automatic mode transitions (Fig.~\ref{fig:cap_removal}b). Initially in vision mode, the sensor identifies the green cap position through color segmentation despite moiré overlay. Upon approach and contact (phase 2), the system switches to tactile mode, where fringe deformation confirms sufficient normal force ($F_z > 0$) for secure grip. During rotation (phase 3), the sensor tracks twist torque $T_z$ through fringe orientation changes, providing feedback for controlled unscrewing. After cap removal (phase 4), the sensor returns to vision mode, confirming task completion through visual verification.

This hand-assisted demonstration validates four capabilities: vision mode for object tracking, tactile mode for force/torque feedback during manipulation, automatic mode switching to eliminate blind spots common in opaque sensors, and the use of the same optical path for both sensing modalities without requiring mechanical reconfiguration. The continuous visual feedback enables error detection and recovery, as the operator can observe both the target and contact state simultaneously.

\subsection{Automated Force-Controlled Manipulation}\label{sec:fixed_base}
We extend the dual-mode capability to automated robotic manipulation, demonstrating a precise force-controlled cap removal with fixed-base bottles. This experiment quantifies MoiréTac's ability to maintain consistent force/torque profiles through an incremental rotation strategy that prevents slippage and cumulative errors.

\begin{figure}[ht]
    \centering
    \includegraphics[width=1\columnwidth]{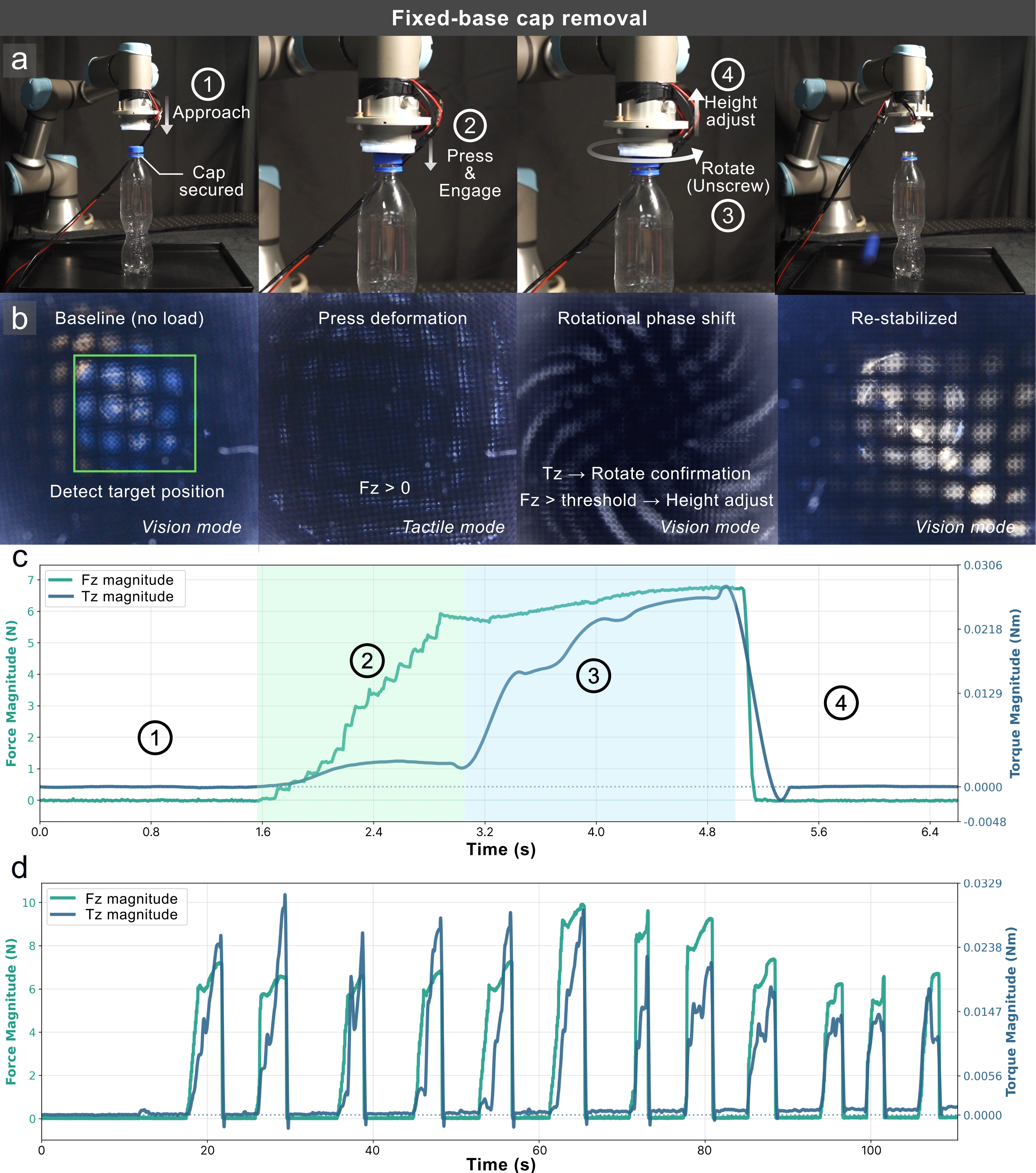}
    \caption{Fixed-base automated cap removal. (a) Robotic execution sequence for each rotation cycle: (1) vision-guided approach, (2) press and engage, (3) controlled partial rotation with height adjustment, (4) release. (b) Sensor views showing vision-to-tactile transitions with force/torque feedback. (c) Single rotation cycle timeline showing coordinated \(F_z\) and \(T_z\) control during phases 1–4 (6.4\,s). (d) Complete cap removal timeline showing 12 incremental rotation over 112\,s.}
    \label{fig:fixed_base}
\end{figure}

The robotic system employs an incremental rotation strategy, repeating a four-phase control cycle until complete cap removal (Fig.~\ref{fig:fixed_base}a,b). This incremental approach prevents slip accumulation that continuous rotation would cause, with each cycle executing a partial rotation under controlled force limits to ensure consistent grip throughout the unscrewing process. For the timeline shown in (Fig.~\ref{fig:fixed_base}c), Phase~1 uses vision mode to align within the cap perimeter. Upon contact detection (\(F_z>0.5\,\mathrm{N}\)), phase~2 gradually increases normal force to \(6\,\mathrm{N}\) for secure grip. During rotation (phase~3), the system executes a controlled partial turn while monitoring \(T_z\) for thread resistance and continuously adjusting height to accommodate the cap's upward displacement as threads disengage. Phase~4 releases the cap, allowing vision mode to verify rotation progress before the next cycle.

The complete removal process (Fig.~\ref{fig:fixed_base}d) requires approximately 12 rotation cycles over 112\,s, successfully demonstrated in repeated trials ($N$=30). Each peak corresponds to one grip–rotate–release sequence, with consistent force profiles: mean peak \(F_z = 7.2 \pm 0.8\,\mathrm{N}\) and mean peak \(T_z = 0.024 \pm 0.003\,\mathrm{Nm}\) per cycle. Early cycles show higher torque resistance as threads are fully engaged, while later cycles exhibit reduced resistance as the cap loosens. Cycle~5 exhibits the highest peaks (\(10\,\mathrm{N}\), \(0.032\,\mathrm{Nm}\)), likely due to thread binding, yet the system maintains control through force-limited operation.

This incremental approach offers advantages including: (i) prevents accumulated slip errors through periodic re-gripping, (ii) enables visual verification between cycles, (iii) maintains target grip force without sensor drift, and (iv) adapts to varying thread resistance throughout removal. The consistent force/torque ratio \((T_z/F_z \approx 0.003\,\mathrm{m})\) across cycles indicates low cross-talk between sensing axes. Vision mode re-engagement between cycles (visible as brief returns toward baseline) provides continuous task monitoring. This automated demonstration confirms MoiréTac provides the sensitivity, dynamic range, and dual-mode capability suitable for dexterous manipulation requiring coordinated force–torque control.



\section{Conclusions}\label{sec:conclusions}

In this paper, we present MoiréTac, a visuotactile sensor that leverages moiré pattern amplification to achieve continuous force/torque sensing while preserving optical transparency. The main contribution lies in establishing analytical relationships between interference patterns and mechanical loads, providing a physics-grounded framework that connects observable features (intensity, phase gradient, orientation, period) to 6-axis forces and torques through optical principles.

Experimental validation confirmed the sensor's performance across multiple modalities. Force/torque characterization achieved \(R^2>0.98\) across all axes, with high accuracy for twist measurement (\(R^2=0.99\)) due to direct fringe-rotation coupling. Sensitivity-tuning experiments demonstrated nearly threefold gain adjustment through geometric design parameters. The transparent architecture enabled object recognition despite moiré overlay, while manipulation experiments showed completed force-controlled cap removal with consistent profiles (\(7.2 \pm 0.8\,\mathrm{N}\)) maintained via an incremental rotation strategy.

MoiréTac's 45 mm diameter and rigid substrate are optimized for parallel-jaw grippers, but alternative form factors could suit curved or miniaturized end-effectors. The moiré overlay sacrifices optical clarity for texture, which can be adjusted for clarity-focused applications. Future work could explore nanoimprinted gratings for flexible sensors, high-frequency phase analysis for slip prediction, and physics-informed learning using analytical relationships (Eqs. 1–9) to enhance data efficiency. These developments will expand moiré sensing into specialized domains, while scaling to multi-device production will need to address cross-device variability and environmental robustness (e.g., temperature). 

The moiré interferometric approach offers advantages for robotic manipulation, supporting gradient-based analysis over the entire contact area. Applications could benefit from dual-mode functionality in quality inspection, collaborative tasks with reduced occlusion, and precision assembly, where dense spatial information enables fine control. As robotic systems demand richer sensory feedback, combining force measurement with visual awareness adds value to the visuotactile sensing toolkit.

\end{document}